\documentclass[a4paper,10pt,twocolumn]{article}
\usepackage[utf8]{inputenc}
\usepackage{amsmath,graphicx}
\usepackage{amsfonts}
\usepackage{amssymb}
\usepackage{mathtools}
\usepackage{multirow}
\usepackage[table]{xcolor}
\usepackage{tabularx}
\usepackage[textwidth=1.1\textwidth,textheight=\textheight]{geometry}
\usepackage{sectsty}
\usepackage{cleveref}
\sectionfont{\normalsize}
\subsectionfont{\normalsize}

\title{\LARGE Content Bias in Deep Learning Image Age Approximation: A new Approach Towards better Explainability}
\author{\normalsize Robert Jöchl and Andreas Uhl\\\normalsize University of Salzburg, Department of Artificial Intelligence and Human Interfaces, \\\normalsize Salzburg, Austria\\
\ \normalsize \{robert.joechl, andreas.uhl\}@plus.ac.at}
\date{}

\begin{document}

\maketitle

\begin{abstract}
In the context of temporal image forensics, it is not evident that a neural network, trained on images from different time-slots (classes), exploits solely image age related features. Usually, images taken in close temporal proximity (\textit{e.g.}, belonging to the same age class) share some common content properties. Such content bias can be exploited by a neural network. In this work, a novel approach is proposed that evaluates the influence of image content. This approach is verified using synthetic images (where content bias can be ruled out) with an age signal embedded. Based on the proposed approach, it is shown that a deep learning approach proposed in the context of age classification is most likely highly dependent on the image content. As a possible countermeasure, two different models from the field of image steganalysis, along with three different preprocessing techniques to increase the signal-to-noise ratio (age signal to image content), are evaluated using the proposed method.
\end{abstract}

\section{Introduction}
\label{sec:intro}
In temporal image forensics, the assumed situation is that a forensics analyst is provided with a set of chronologically ordered trusted images and a second not trustworthy set (from the same imager). The main objective is to approximate the age of images from the not trustworthy set relative to the trusted set. This age estimation is usually based on time-dependent traces that are hidden in a digital image. The most common and well-known traces in this context are in-field sensor defects \cite{leung2009}. In-field sensor defects accumulate over time. Thus, the age of an image (relative to images from the same imager) can be approximated based on the detected defects present. Two methods that explicitly exploit the presence of in-field sensor defects were proposed in \cite{Fridrich11a} and \cite{joechl20a}. Unlike these traditional techniques, a Convolutional Neural Network (CNN) learns the classification features used independently. Ahmed et al. utilized two well-known CNN architectures for age classification (\textit{i.e.}, the AlexNet \cite{krizhevsky12} and GoogLeNet \cite{szegedy15a}) in \cite{ahmed20a}. The best performing model was the AlexNet in the transfer learning mode with a reported accuracy of up to 88\% for a five-class classification problem. In \cite{Joechl21b}, we systematically investigated the influence of the presence of strong in-field sensor defects when training a CNN for age classification. Considering the investigated `five-crop-fusion' scenario (where five networks are trained on different fixed image patches each) the presence of a strong in-field sensor defect turns out to be irrelevant for improving the age classification accuracy. For this reason, we suggested in \cite{Joechl21b} that other `age' traces are exploited by the network.

Following the results in \cite{Joechl21b}, we analyzed the device (in)dependence of the learned `age' features in \cite{Joechl22a}. However, based on the results obtained, the question of device (in)dependence could not be answered. In contrast, the reported results cast doubt on whether the inference was based solely on age related features. Deep neural networks can be considered as a `black box'. For example, in the context of deep learning image age approximation, it is not evident that inference is based solely on detected age traces. In principle, it is likely that images taken in close temporal proximity (\textit{e.g.}, belonging to the same age class) share some common content properties (\textit{e.g.}, common scene properties, weather conditions or seasonal commonalities). Such non-age-related correlations can be exploited by the CNN to discriminate between the age classes. The features learned by a standard CNN trained in the context of image age approximation were analyzed in \cite{Joechl23a}. For this purpose, we used a method from the field of explainable artificial intelligence (XAI), \textit{i.e.}, Class Activation Maps (CAMs). Based on this analysis, we showed that most likely the image content (\textit{e.g.,} scene properties and light conditions) is more important for classification than the inherent age signal. However, the analysis of CAMs can become cumbersome when the influence of image content needs to be evaluated for different models.

This work is an extension of \cite{Joechl22a} and \cite{Joechl23a}. We deepen the analysis of the role of image content in the context of feature-learning based image age approximation algorithms (\textit{i.e.}, deep neural networks). For this purpose, the contribution of this work are as follows:
\begin{enumerate}
 \item A novel approach in the context of XAI and deep learning image age approximation is proposed, that evaluates the influence of image content.
 \item The functionality of the proposed XAI method is verified using synthetic images (where content bias can be ruled out) with an age signal embedded.
 \item Based on the proposed XAI method, the influence of image content on different CNN models trained in the context of age classification is evaluated.
\end{enumerate}

The remainder of this paper is organized as follows: the proposed XAI approach is described in section \ref{sec:xai}. In section \ref{sec:exp}, the conducted experiments (\textit{i.e.}, the evaluated models, the used dataset, etc.) are described and the results discussed. The key insights are summarized in the last section \ref{sec:con}.

\section{Explainable Artificial Intelligence (XAI)}
\label{sec:xai}
CNNs usually consist of millions of parameters, which turn the models into a `black box’. The field of XAI focuses on understanding and interpreting the decisions of such deep neural networks. A comprehensive survey of methods in the field of XAI is given in \cite{Arrieta20a,linardatos21,Ivanovs2021a}. In the context of deep learning image age approximation, explainability could consist of understanding the age features exploited by the network. However, the first step is to assess whether image age features are actually being used. We are currently not aware of any existing XAI method in this context. In this work, a novel model agnostic approach is proposed that explains if the decision made is based on comprehensible evidence (\textit{i.e.}, image age features) or on content bias. This approach is a post-hoc method that aims to explain the behavior of a trained model at testing time. The fundamental idea is to present the model with particularly prepared input such that the model output allows to derive if actual age traces are exploited or the model takes advantage of content (bias). For example, data input can be prepared where the age signal is still present, but the content is strongly suppressed. Then the age classification accuracy of the model based on this input is evaluated. If the age classification accuracy is similar to the accuracy achieved based on the original input (as used for training and testing), the image content is probably less important. In an average image (\textit{e.g.}, generated from all images of a certain age class) image content is suppressed to a certain extent, but not entirely so.

\subsection{Average Images}
\label{sec:avg_I}
\begin{figure}
 \centering
 \includegraphics[width=0.22\textwidth]{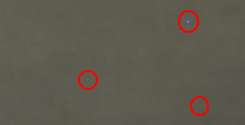}
 \caption{Example of an average image, age traces like in-field sensor defects are preserved.}
 \label{fig:def_example}
\end{figure}

\begin{figure}[ht!b]
    \centering
    \begin{minipage}[b]{0.22\textwidth}
        \centering
        \includegraphics[width=0.84\textwidth]{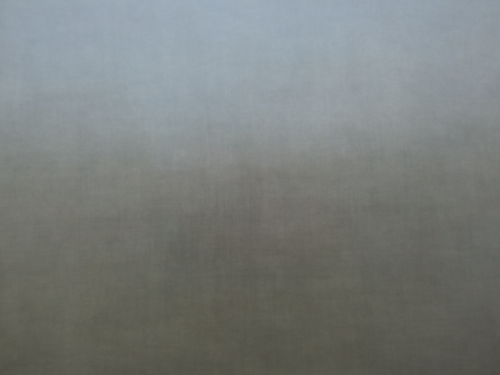}
        \centerline{\small{(a) class 1}}\medskip
    \end{minipage}
    \begin{minipage}[b]{0.22\textwidth}
        \centering
        \includegraphics[width=0.84\textwidth]{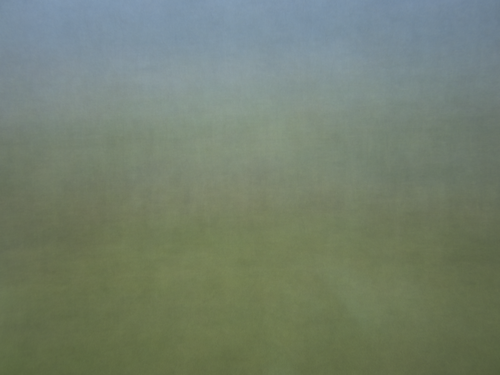}
        \centerline{\small{(b) class 2}}\medskip
    \end{minipage}
    \caption{Example of average images.}
    \label{fig:avg_Is}
\end{figure}

A signal that is stable in terms of position and polarity is basically preserved in an average image, while a random signal is suppressed. Based on the characteristics of known age traces (\textit{i.e.}, in-field sensor defects), the age signal is preserved in average images while the content is suppressed. This is also illustrated in Fig. \ref{fig:def_example}. Thus, if the model has learned to exploit an age signal (\textit{i.e.}, in-field sensor defects), the age classification accuracy of the trained model should be similar when classifying average images or the original input (used for training and testing). However, an average image also represents the average content. As can be seen in Fig. \ref{fig:avg_Is} (a and b), the images of class 2 probably contain more nature scenes (green areas) whereas class 1 images probably contain more urban scenes (more gray than green). These remaining content parts can also be exploited by the model. For this reason, as the core of the XAI approach, an evaluation is proposed on the basis of four different variants of average images.

Before the four variants are described, the assumed situation is defined. Image age approximation can be considered a multi-class classification problem, where the classes are defined by the temporal resolution of the exploited age traces and the available trusted images. For this purpose, let $\theta \in \mathbb{R}^{h \times w}$ be the sum of age traces at a certain point in time that are embedded in an image $Y$ of height $h$ and width $w$. We assume that $\theta$ is constant (except for small variations due to different camera settings) across all images of a given age class $k$ and differs between the other age classes. In contrast to $\theta$, the image content within a given class is more or less random. Furthermore, we assume that the dataset $S$, which is used to train and test the model, is available and that the class definitions are known.

The first variant of average images is a standard average image, \textit{i.e.},
\begin{equation}
  \overline{Y^{k}}(i,j) = \frac{1}{|S_k|} \sum_{Y \in S_k} Y(i,j), 
\end{equation}
where $i$ and $j$ denote the pixel index. Recall, that the age classification accuracy of average images should be similar to the accuracy of the original input $S$ (as used for training and testing) if inference is based (solely) on a hidden age signal.

An image representing the average color of the average image, \textit{i.e.},
\begin{equation}
 \overline{Y^{k}}_{c}(i,j) = \frac{1}{h \times w} \sum_{i^\prime = 1}^{h} \sum_{j^\prime = 1}^{w} \overline{Y^{k}}(i^\prime,j^\prime),
\end{equation}
is considered as second variant. Each pixel in the image has the same value (constant component), thus, any structural components (including the age signal) are eliminated. Since the average color depends strongly on the image content, a successful age classification of $\overline{Y^{k}}_{c}$ would indicate that the image content is exploited by the model.

The structural components, `independent' of color, are considered as the third variant of average images, \textit{i.e.},
\begin{equation}
 \overline{Y^{k}}_{r}(i,j) = \overline{Y^{k}}(i,j) - min(\overline{Y^{k}}).
\end{equation}
To avoid negative values, a subtraction with the minimum value of $\overline{Y^{k}}$ is performed. If $min(\overline{Y^{k}})$ is near zero, the difference to $\overline{Y^{k}}$ is marginal. In terms of present age signal and average content (except color), $\overline{Y^{k}}$ and $\overline{Y^{k}}_{r}$ are identical. Thus, if the inference is based on an age signal, a successful age classification should also be possible with $\overline{Y^{k}}_{r}$.

The last variant is the filtered average image, \textit{i.e}.,
\begin{equation}
 \overline{Y^{k}}_{f}(i,j) = F(\overline{Y^{k}}(i,j)),
\end{equation}
where $F$ represents a median filter with kernel size 5. The median filter filters out all high-frequency image components (\textit{i.e.}, including in-field sensor defects). Thus, if inference is based on such components a successful age classification of $\overline{Y^{k}}_{f}$ should no longer be possible. Average content and age signal present in low-frequency image components are still present.

\begin{figure}[ht!b]
    \centering
    \begin{minipage}[b]{0.11\textwidth}
        \centering
        \includegraphics[width=0.98\textwidth]{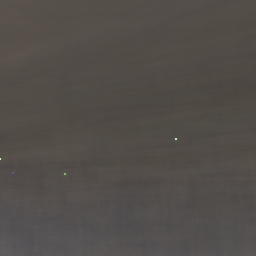}
        \centerline{\small{$\overline{Y^{k}}_{}$}}\medskip
    \end{minipage}
    \begin{minipage}[b]{0.11\textwidth}
        \centering
        \includegraphics[width=0.98\textwidth]{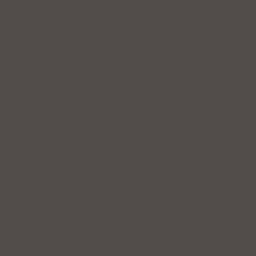}
        \centerline{\small{$\overline{Y^{k}}_{c}$ }}\medskip
    \end{minipage}
    \begin{minipage}[b]{0.11\textwidth}
        \centering
        \includegraphics[width=0.98\textwidth]{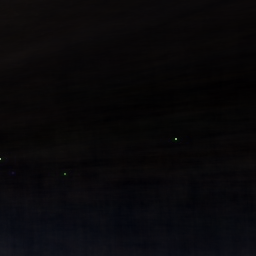}
        \centerline{\small{$\overline{Y^{k}}_{r}$}}\medskip
    \end{minipage}
    \begin{minipage}[b]{0.108\textwidth}
        \centering
        \includegraphics[width=0.98\textwidth]{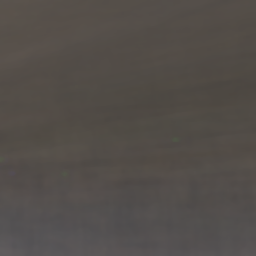}
        \centerline{\small{$\overline{Y^{k}}_{f}$}}\medskip
    \end{minipage}
    \caption{Example of the different variants generated.}
    \label{fig:avg_I_examples}
\end{figure}
An example of all generated variants is given in Fig. \ref{fig:avg_I_examples}. In summary, if the trained model has learned to exploit (solely) an age signal, the age classification accuracy of $\overline{Y^{k}}$ and $\overline{Y^{k}}_{r}$ should be similar to the accuracy achieved when the original input $S$ (used for training and testing) is provided. In contrast, a similar accuracy based on $\overline{Y^{k}}_{c}$ would indicate that content is exploited. $\overline{Y^{k}}_{f}$ evaluates if a signal from the high-frequency image components is exploited. For example, if in-field sensor defects are exploited by the model, successful age classification based on $\overline{Y^{k}}_{f}$ should not be possible.

\subsection{Age Signal Embedded in Synthetic Images}
\label{sec:as}
Synthetic images are computer generated (rendered) and do not contain any artifacts or noise introduced by the image acquisition pipeline (\textit{i.e.}, $Y_{SYN} = I $). Hence, they do not exhibit an age signal. To evaluate the functionality of the proposed XAI method, the method is applied to a trained model where content bias can be ruled out. For this purpose, an age signal is embedded into synthetic images. Thus, truly identical images (in terms of image content) are available per class, with only a different version of $\theta$ being embedded. If a model is trained on these images, the image content can be definitely ruled out and only the embedded age signal is exploited.

For this work, a subset (400 rendered images) of the Grand Theft Auto 5 dataset \cite{Richter16a} is considered per class. A series of dark field images (DFI) taken with two of our own devices at two different points in time (with a time difference of about 4 years) is used to estimate the age signal that is embedded in the rendered images. DFIs are calibration images where the shutter of the camera is closed and thus the incident light is set to zero (\textit{i.e}. $I=0$).

In \cite{Fridrich11a}, Fridrich and Goljan defined the following output model of a sensor:
\begin{equation}
  Y = I + IK + \tau D + c + \Theta,
\end{equation}
where $K$ denotes the Photo-response non-uniformity (PRNU), $D$ the dark current, $c$ a fixed offset and $\Theta$ is a collection of all other noise sources. Since $D$ depends on exposure time, ISO setting, and temperature, these factors are combined into $\tau$. In-field sensor defects usually have a high value of $D$ and/or $c$. Since the incident light is set to zero (\textit{i.e}. $I=0$) when capturing DFIs, 
\begin{equation}
  Y_{DFI} = \tau D + c + \Theta.
\end{equation}
Thus, only additive parameters (\textit{e.g.}, in-field sensor defects) remain. $\Theta$ could also contain an (unknown) age signal that is also retained. However, random parts in $\Theta$ are mitigated through averaging a series of DFIs. Since only additive parameter remain, the estimated age signal $\hat{\theta}_k$ of a certain class $k$ is simply added to $Y^{SYN}$, \textit{i.e.}, 
\begin{equation}
 Y_{SYN}^{k} = I + \hat{\theta}^k = I + \overline{\tau D}^k + \overline{c}^k + \overline{\Theta}^k.
\end{equation}
\begin{figure}[ht!b]
    \centering
    \begin{minipage}[b]{0.22\textwidth}
        \centering
        \includegraphics[width=0.88\textwidth]{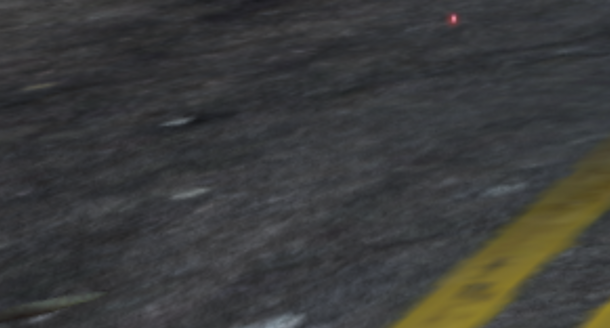}
        \centerline{\small{(a) example $Y_{SYN}^1$}}\medskip
    \end{minipage}
    \begin{minipage}[b]{0.22\textwidth}
        \centering
        \includegraphics[width=0.88\textwidth]{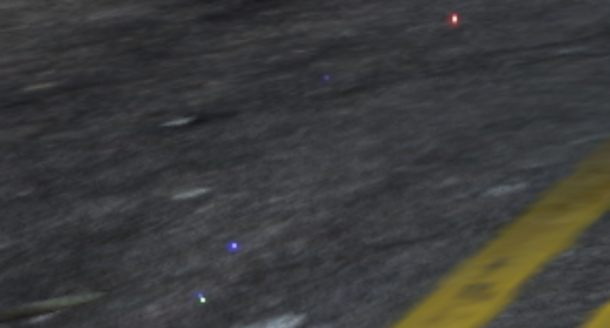}
        \centerline{\small{(b) example $Y_{SYN}^2$}}\medskip
    \end{minipage}
    \begin{minipage}[b]{0.22\textwidth}
        \centering
        \includegraphics[width=0.78\textwidth]{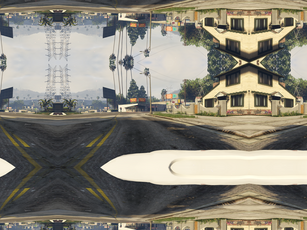}
        \centerline{\small{(c) example mirroring}}\medskip
    \end{minipage}
    \caption{Example of synthetic images.}
    \label{fig:syn_imgs_example}
\end{figure}
Thus, all age traces that depend on $I$ are not considered.

An example of the embedded age signal (in-field sensor defects) is illustrated in Fig. \ref{fig:syn_imgs_example} (a and b). The dimension of $\hat{\theta}$ ($w \times h$) corresponds to the pixel resolution of the camera used to capture the DFIs. However, this resolution might not correspond to the resolution of the synthetic images. To obtain a resolution corresponding to $\hat{\theta}$, $Y_{SYN}$ is expanded by mirroring the boundary regions (see Fig. \ref{fig:syn_imgs_example} (c)). Recall, that the evaluation of models trained on these synthetic data is just to validate the functionality of the proposed XAI method. The actual evaluation of whether the image content or the age signal is learned is conducted on regular scene images.

\section{Experiments}
\label{sec:exp}
To evaluate the influence of image content, five different models are trained in the context of image age classification. Before these models are trained on regular scene images, these models are trained on synthetic images (with an age signal embedded) to verify the functionality of the proposed XAI method.

\subsection{Evaluated Models}
The first model considered is one that has already been used for image age classification, \textit{i.e.}, the AlexNet in transfer learning mode proposed in \cite{ahmed20a}. In \cite{ahmed20a}, an image was cropped into 48 non-overlapping blocks of size $500 \times 500$ and each block was fed separately through the network. The final class prediction was determined by a majority vote of each block's predictions. It is not clearly described how the 48 non-overlapping blocks used are extracted, or how many epochs were trained with which batch size. In this work, the first 48 adjacent blocks, starting in the top left corner of the image, are extracted. When applying transfer learning over 60 epochs with a batch size of 4, the results reported in \cite{ahmed20a} can be confirmed. The AdaMax is used as an optimizer with a learning rate (lr) of 0.0001.

Age traces can be interpreted as a weak signal embedded in an image. The field of image steganalysis\cite{DeRosal23a} is the science of detecting whether there is a secret message (weak signal) hidden in an image. For this reason, the methods proposed in the field of image steganalysis have been suggested to be suited to detect subtle age traces \cite{Joechl21b}. The Steganalysis Residual Network (SRNet) \cite{boroumand18a} is a model that was proposed in the field of image steganalysis and has already been used in the context of image age approximation \cite{Joechl21b,Joechl22a,Joechl23a}. The key part of the SRNet is the first seven layers, where no pooling operation is involved. Since pooling acts as a low-pass filter, omitting it avoids suppressing the noise-like stego (age) signal. In \cite{Joechl21b}, the SRNet was trained using different learning scenarios to investigate the influence of the presence of strong in-field sensor defects. The `five-crop-fusion’ learning scenario was the most position dependent scenario. In particular, five different SRNets are trained on patches ($256 \times 256$) taken from a fixed position in the images. These positions are at the top left (`tl'), the top right (`tr'), the bottom left (`bl'), the bottom right (`br') corners and in the center of the image (`ce'). The final class prediction is obtained by fusing together the different outputs of the individual sub-models. As already shown in \cite{Joechl23a}, when a `standard' SRNet is trained on regular scene images, inference is most likely based on image content.

One way to reduce the interference of the image content \textbf{during training} is to preprocess the input data. For this purpose, a possibility is to apply a denoising filter $F$, and then obtain the residuals by subtracting the filtered image from the original image, \textit{i.e.}, $R = Y - F(Y)$. By considering a pixel's dynamic range from 0 to 255, a single residual value $R_i$ can have a value $R_i \in \mathbb{Z}: -255 \leq R_i \leq 255$. For this work, a median filter with kernel 3 is used as denoising filter $F$. To force the residuals to be in the same range as the pixel values, the absolute values are considered (\textit{i.e.}, $R = |Y - F(Y)|$). Only high-frequency image components (\textit{e.g.}, edges) are present in median filter residuals. Thus, a lot of content is suppressed while the age signal is preserved (i.e., the parts contained in the high-frequency image components, such as in-field sensor defects). This increases the signal-to-noise ratio (age signal to image content), which should help the model to learn age traces and is considered as third model (SRNet-res).

In \cite{Bayar18a}, Bayar et al. proposed a preprocessing layer (the constrained convolutional layer) that should learn residuals. To learn residuals, the following constraints
\begin{equation}
    \begin{cases}
    w_{k}(0,0) = -1\\
    \sum_{m,n \neq 0} w_{k} (m,n) = 1
    \end{cases}
\end{equation}
are applied to each of the $K$ filters in the constraint convolution layer, where $k$ denotes the $k^{th}$ filter and the central weight $w$ is denoted by spatial index (0,0). Since an RGB input is considered in this work and the magnitude of the age signal may vary between color channels, the described constraint is applied to each layer of the filter kernel. The SRNet equipped with constrained convolutional layer (content suppression layer that increases the signal-to-noise ratio) is considered as fourth model and is denoted by SRNet-cs. All SRNet variants (SRNet, SRNet-res and SRNet-cs) are trained in the `five-crop-fusion’ approach over a fixed number of 180 epochs with a batch size of 4 as done in \cite{Joechl21b,Joechl22a,Joechl23a}. The lr of 0.001 is decreased to 0.0001 after 100 epochs and the AdaMax is used as an optimizer.

The last model is again a model proposed in the context of image steganalysis, the ZhuNet\footnote{https://github.com/1204BUPT/Zhu-Net-image-steganalysis} \cite{ZhangR20a}. The ZhuNet already includes a preprocessing layer. In particular, 30 basic high-pass filters of the Spatial Rich Model (SRM) \cite{Fridrich12a} were used to extract residuals. This preprocessing layer can also be considered as content suppression layer that increases the signal-to-noise ratio. The parameters of the preprocessing layer (SRM filters) are not updated during training. For the ZhuNet, the defined optimizer is Stochastic Gradient Descend with a lr of 0.001 and momentum 0.95. The lr is divided by 5 after 20, 35, 50 and 65 epochs. A total of 80 epochs are conducted using the `five-crop-fusion' approach. For synthetic images, the ZhuNet is trained over 240 epochs, as it takes longer to converge. For all five models, cross entropy is used as loss type. A comparison of SRNet and ZhuNet is given in \cite{ZhangR20a}.

Recall, that the main contribution of this work is not to improve or compare the age classification results. Rather, it is to assess whether models that have learned to `successfully' discriminate between age classes utilize age traces or image content and to highlight the problem of content bias.

\subsection{Dataset}
The Northumbria Temporal Image Forensics (NTIF) \cite{NTIFDB} database is a publicly available dataset for temporal image forensics. The NTIF dataset comprises images from 10 different cameras: two Canon IXUS115HS (Nc01 \& Nc02), two Fujifilm S2950 (Nf01 \& Nf02), two Nikon Coolpix L330 (Nn01 \& Nn02), two Panasonic DMC TZ20 (Np01 \& Np02) and two Samsung pl120 (Ns01 \& Ns02). For each device, approximately 71 time-slots ranging over 94 weeks (between 2014 and 2016) are available. Further details about the dataset can be found in \cite{NTIFDB}. Images from this database are also used in other image forensics related publications, \textit{i.e.}, \cite{alani2017, lawgaly2017}.

In the context of image age approximation, the NTIF database is used in \cite{ahmed20a,Joechl22a,Joechl23a, ahmed21a}. The authors in \cite{ahmed20a} divided the first 25 time-slots into five age classes. As already mentioned in the introduction, an overall classification accuracy of up to 88\% is reported for these five classes (by AlexNet with transfer learning mode). To train the AlexNet, the same class definitions as in \cite{ahmed20a} are used (5 age classes). For training the SRNet variants and the ZhuNet only the first and last classes (\textit{i.e.,} time-slot 1-5 and 21-25, respectively) are considered (binary classification problem as done in \cite{Joechl22a,Joechl23a}). There is a time difference of about 4 months between time-slot 1-5 and 21-25. For training the SRNet variants and the ZhuNet, images from two of our own imagers, a Canon PowerShotA720IS (Pc01) and a Nikon E7600 (Pn01), are used additionally. For both imagers, two classes are considered, with a time difference of about 12 and 15 years, respectively. Images from these two devices are also used in \cite{Joechl22a,Joechl23a,joechl20a,Joechl21b}. The two sets of DFIs used to generate the synthetic images were also acquired with these two imagers. However, these two sets were not taken at the same time as the regular scene images.

Between the first and second class of Pc01, a total of 27 new `strong' in-field sensor defects have developed, 3 of which are located in the five-crop regions. These `strong' defects are detected by thresholding the median filter residuals of the generated average images. For all other imagers, no `strong' in-field sensor defects are detected in the five-crop regions. It is also noticeable that the camera parameters (ISO, exposure time and focal length) differ significantly between the classes (\textit{e.g.}, the mean ISO/exp.time/focal length for class 1 and 2 of Nc02 are 342/0.013/7.45 and 163/0.0076/5 with a standard deviation of 360/0.020/4.84 and 89/0.011/0), which changes the image characteristics (\textit{i.e.}, a higher ISO/exp.time lead to more noise, while different focal lengths can lead to different distortions). This is also an indication of different lighting conditions between the two classes.

In \cite{ahmed20a}, the authors used 60\% of samples per class for training the AlexNet and 40\% for testing. The same sampling ratio is applied in this work as well for training the AlexNet. However, for the remaining models (SRNet variants and ZhuNet) the sampling ratio as used in \cite{Joechl22a,Joechl23a} is considered, \textit{i.e.}, 80\% of the images per class are selected as train set and the remaining 20\% are divided equally between a validation set and a test set. The evaluation of all five considered models is performed 8 times for each imager, with samples selected randomly and independently.

To evaluate a trained model based on the proposed XAI approach, 20 different sets of average images are generated per class, with 80\% of all samples (including training, validation and testing data) randomly selected for each set. However, 80\% of the samples are used only for the class with fewer samples. The other class is undersampled accordingly. This is to ensure that a balanced set is used to generate the average images. If preprocessing is applied (\textit{i.e.}, SRNet-res), it is always applied (\textit{i.e.}, to the training, validation and test set, as well as before creating the average images).

\subsection{Evaluation Metric}
As described in section \ref{sec:avg_I}, the age signal in average images should be preserved while content is suppressed. For this reason, the classification accuracy of the trained model (when inference is based on an age signal) for average images should be similar to that for original inputs $S$ (\textit{i.e.}, the images used to train the model, like regular scene images or median filter residuals). Similarity can be measured by the difference between two quantities. Hence, we compute the mean difference of accuracy values ($\overline{\Delta}$) resulting from providing average images ($acc^{\overline{Y}}$) and original inputs ($acc^{S}$), \textit{i.e.}, 
\begin{equation}
 \overline{\Delta}(\overline{Y})= \frac{1}{N} \sum_{i=1}^{N} (acc_{i}^{S} - 0.5) - |acc_{i}^{\overline{Y}} - 0.5|,
 \label{eq:diff}
\end{equation}
where $N$ denotes the number of runs. In the context of average images, the accuracy $acc^{\overline{Y}}$ is considered as a measure of separability, where $acc^{\overline{Y}} = 0$ and $acc^{\overline{Y}} = 1$ indicates a perfect separation of the average images. Since the average images are balanced, an accuracy of 0.5 represents the worst case in terms of separability (in a binary classification problem). For this reason, an accuracy value of, for example, 0.2 and 0.8 can be considered equivalent in terms of separability. In equation (\ref{eq:diff}), this is expressed by the term $|acc_{i}^{\overline{Y}} - 0.5|$. An accuracy of less than 0.5 may occur when providing average images because the pixel value distribution of the average images differs from the distribution of the images (regular images or median filter residuals) used for training. The term $(acc_{i}^{S} - 0.5)$ is used to bring the accuracy obtained by the original inputs into the same range. However, equation (\ref{eq:diff}) can only be applied in a binary classification scenario. A generalized version independent of the number of classes is to simply compute the mean difference of accuracy values resulting from providing original inputs and average images, \textit{i.e.},
\begin{equation}
 \overline{\Delta}_{GEN}(\overline{Y})= \frac{1}{N} \sum_{i=1}^{N} acc_{i}^{S} - acc_{i}^{\overline{Y}}.
 \label{eq:diff_gen}
\end{equation}

For both variants (equation (\ref{eq:diff}) and (\ref{eq:diff_gen})) it is expected that, if the model has learned to exploit an age signal, $\overline{\Delta}(\overline{Y})$ and $\overline{\Delta}(\overline{Y}_{r})$ should be approximately zero. In contrast, $\overline{\Delta}(\overline{Y}_{c}) \approx 0$ would indicate that most likely content is exploited, since $\overline{Y}_{c}$ is strongly dependent on image content. If age traces in high-frequency image components (\textit{e.g.}, in-field sensor defects) are exploited, $\overline{\Delta}(\overline{Y}_{f})$ should be clearly greater than zero. In addition, an $\overline{\Delta}(\overline{Y}) < 0$ indicates that the accuracy based on the average images is better than based on the original input. However, since a neural network can be considered a `black box', there is no exact threshold for $\overline{\Delta}(.)$ that rules out exploited content bias.

\subsection{Experimental Results}
\begin{figure}[ht!b]
    \centering
    \includegraphics[width=0.44\textwidth]{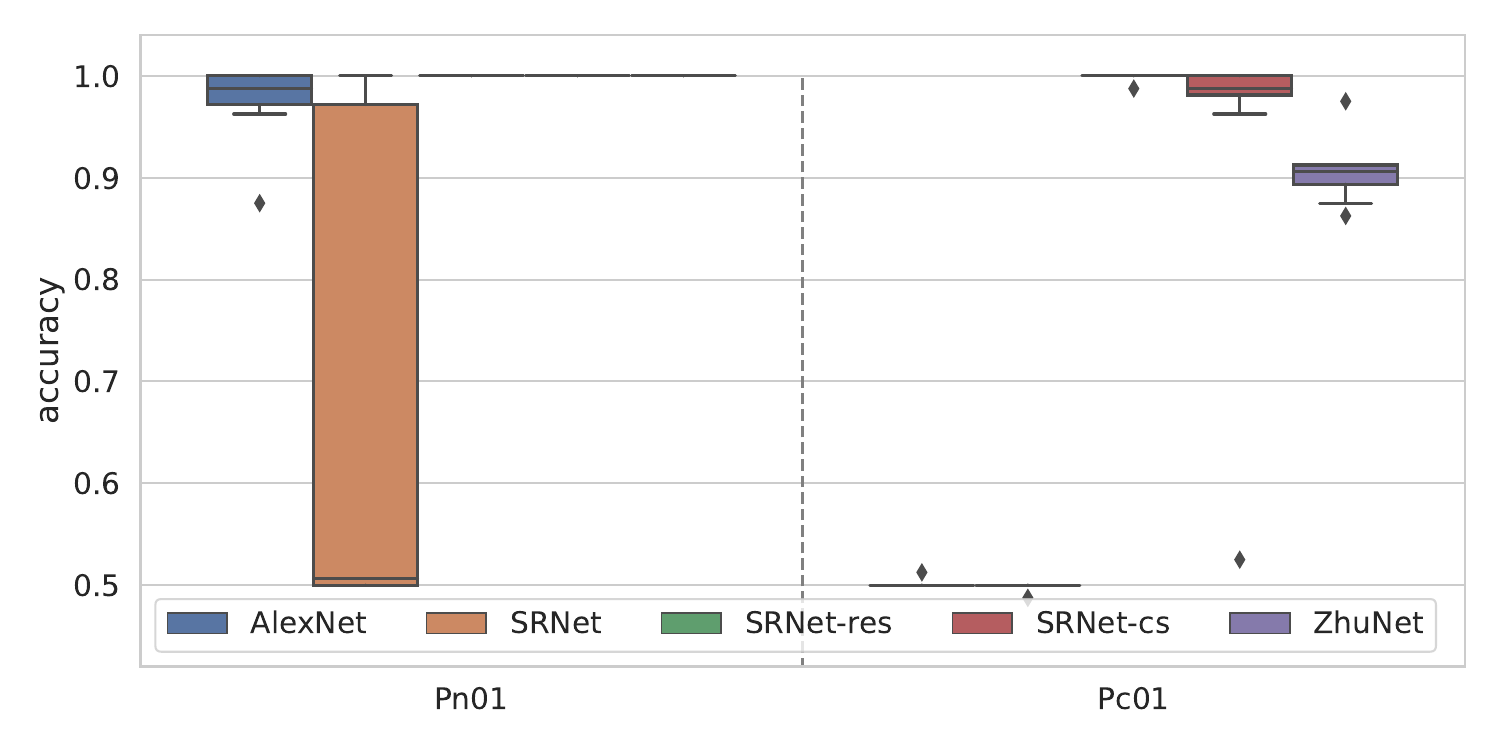}
    \caption{Classification accuracy, when the models are trained on syn. images. The model order corresponds to the legend.}
    \label{fig:exp_results_syn}
\end{figure}

Before analyzing the results obtained when applying the proposed XAI method, Fig. \ref{fig:exp_results_syn} illustrates the classification accuracy obtained (on the test set) when the models are trained on synthetic images. It can be clearly seen that the age signal extracted from Pn01 is better detectable. All models that include some form of preprocessing to increase the signal-to-noise ratio (\textit{i.e.}, SRNet-res, SRNet-cs or ZhuNet) perform significantly better than without (\textit{i.e.}, AlexNet and SRNet). This already indicates that it is difficult for a `standard' CNN trained on raw pixel values to exploit the subtle age traces, even if distractions from the image content can be ruled out.

Table \ref{tab:syn_results} shows the results obtained when applying the proposed XAI method. As expected, if age classification is possible based on original input (\textit{i.e.}, $\overline{acc^{S}} > 0.5$), age classification based on average images ($\overline{Y}$) and the structural components of the average images ($\overline{Y}_{r}$) is equally possible (\textit{i.e.}, $\overline{\Delta}(.) \approx 0$). However, if median filter residuals are used for training (SRNet-res, Table \ref{tab:syn_results}), $\overline{Y}_{r}$ seems to yield more stable results. Age classification based on average color ($\overline{Y}_{c}$) and the filtered average image ($\overline{Y}_{f}$) is not possible (\textit{i.e.},  $\overline{acc^{S}} - \overline{\Delta}(.) \approx 0.5$). This is reasonable since most likely in-field sensor defects are exploited by the network. Based on the results, it can be stated that the proposed XAI method works as expected.

\begin{table}[htb]
\centering
\caption{$\overline{\Delta}(\cdot)/\overline{\Delta}_{GEN}(\cdot)$, models trained on syn. imgs (2 classes). Recall: An age signal is present in $\overline{Y}$ and $\overline{Y}_{r}$ and not in $\overline{Y}_{c}$. Expectation if an age signal is learned, $\downarrow : \overline{\Delta} \approx 0$ and $\uparrow : \overline{\Delta} >> 0$. $\overline{Y}_{f}$, high frequency image components (e.g., defects) are filtered out.}
\scriptsize
\label{tab:syn_results}
\begin{tabular}{l|c|c|c|c|c}\multicolumn{6}{c}{}\\ [-0.4em]
\multicolumn{6}{c}{\textbf{AlexNet}} \\
IDs & $\overline{Y} \downarrow$ & $\overline{Y}_{c} \uparrow$ & $\overline{Y}_{r} \downarrow$ & $\overline{Y}_{f}$& $\overline{acc^{S}}$ \\ \hline
Pn01 & -0.02/-0.02 & 0.48/0.48 & -0.02/-0.02 & 0.48/0.48 & 0.98 \\
Pc01 & -0.03/-0.03 & 0.02/0.02 & -0.02/-0.02 & 0.00/0.00 & 0.54 \\ \hline \hline
\multicolumn{1}{c|}{$\varnothing$}& -0.02/-0.02& 0.25/0.25& -0.02/-0.02& 0.24/0.24& 0.76 \\ 
\multicolumn{6}{c}{\textbf{SRNet}} \\
IDs & $\overline{Y} \downarrow$ & $\overline{Y}_{c} \uparrow$ & $\overline{Y}_{r} \downarrow$ & $\overline{Y}_{f}$& $\overline{acc^{S}}$ \\ \hline
Pn01 & -0.02/-0.02 & 0.18/0.18 & 0.00/0.00 & 0.18/0.18 & 0.68\\
Pc01 & 0.00/0.00 & 0.00/0.00 & 0.00/0.00 & 0.00/0.00 & 0.50 \\ \hline \hline
\multicolumn{1}{c|}{$\varnothing$}&-0.01/-0.01 & 0.09/0.09 & 0.00/0.00 & 0.09/0.09 & 0.59 \\
\multicolumn{6}{c}{\textbf{SRNet-res}} \\
IDs & $\overline{Y} \downarrow$ & $\overline{Y}_{c} \uparrow$ & $\overline{Y}_{r} \downarrow$ & $\overline{Y}_{f}$& $\overline{acc^{S}}$ \\ \hline
Pn01 & 0.00/0.00 & 0.50/0.50 & 0.00/0.00 & 0.50/0.50 & 1.00\\
Pc01 & 0.06/0.06 & 0.50/0.50 & 0.00/0.00 & 0.50/0.50 & 1.00 \\ \hline \hline
\multicolumn{1}{c|}{$\varnothing$}& 0.03/0.03 & 0.50/0.50 & 0.00/0.00 & 0.50/0.50 & 1.00 \\
\multicolumn{6}{c}{\textbf{SRNet-cs}} \\
IDs & $\overline{Y} \downarrow$ & $\overline{Y}_{c} \uparrow$ & $\overline{Y}_{r} \downarrow$ & $\overline{Y}_{f}$& $\overline{acc^{S}}$ \\ \hline
Pn01 & 0.00/0.00 & 0.50/0.50 & 0.00/0.00 & 0.50/0.50 & 1.00\\
Pc01 & 0.00/0.00 & 0.43/0.43 & 0.00/0.00 & 0.42/0.42 & 0.93\\ \hline \hline
\multicolumn{1}{c|}{$\varnothing$}& 0.00/0.00 & 0.47/0.47 & 0.00/0.00 & 0.46/0.46 & 0.97 \\
\multicolumn{6}{c}{\textbf{ZhuNet}} \\
IDs & $\overline{Y} \downarrow$ & $\overline{Y}_{c} \uparrow$ & $\overline{Y}_{r} \downarrow$ & $\overline{Y}_{f}$& $\overline{acc^{S}}$ \\ \hline
Pn01 & -0.01/-0.01 & 0.49/0.49 & -0.01/-0.01 & 0.49/0.49 & 0.99 \\
Pc01 & -0.05/-0.05 & 0.44/0.44 & -0.05/-0.05 & 0.36/0.36 & 0.94 \\ \hline \hline
\multicolumn{1}{c|}{$\varnothing$}& -0.03/-0.03& 0.47/0.47& -0.03/-0.03& 0.42/0.42& 0.97 \\
\end{tabular}
\end{table}

\begin{figure}[ht!b]
    \centering
    \includegraphics[width=0.48\textwidth]{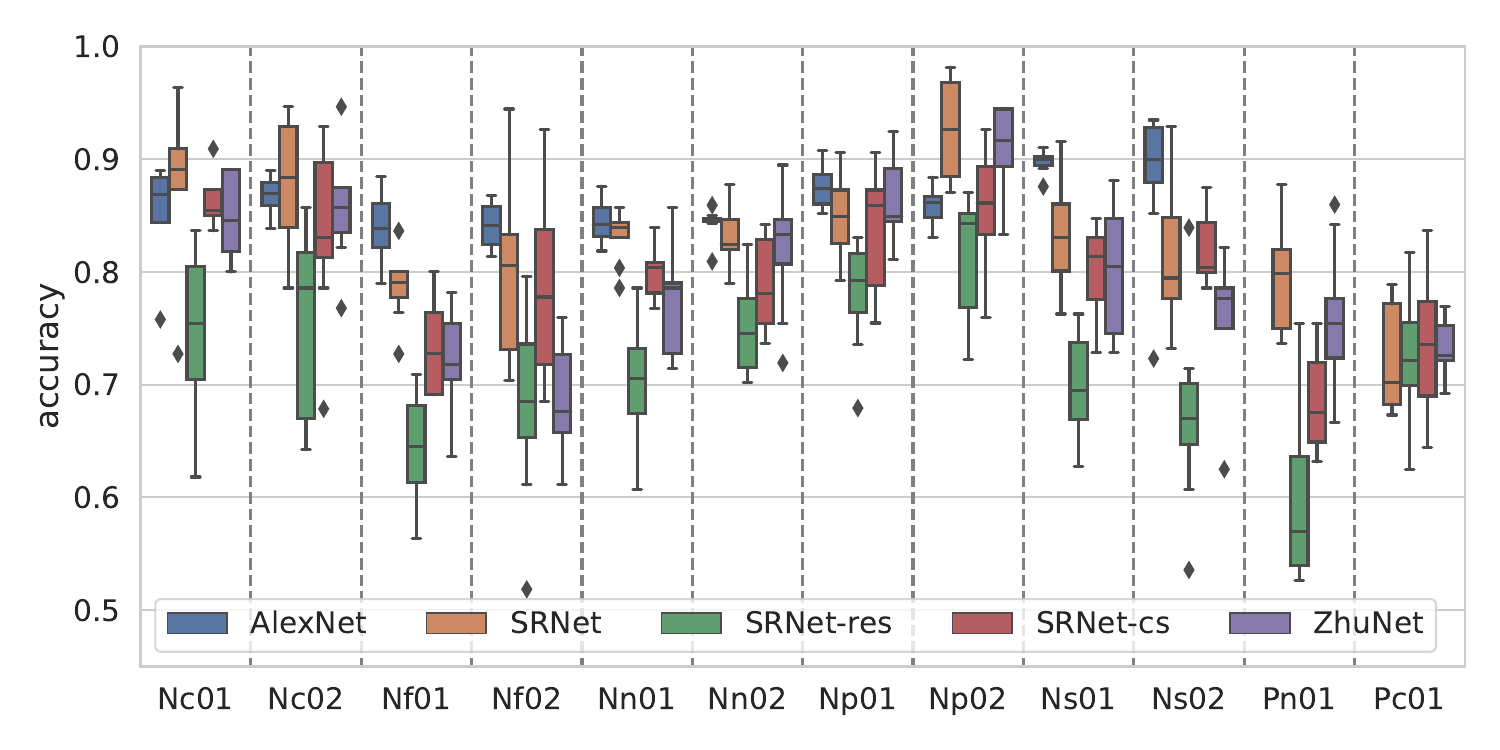}
    \caption{Classification accuracy, when the models are trained on regular images (AlexNet 5 classes, all other 2 classes).}
    \label{fig:exp_results_reg}
\end{figure}

\begin{table}[htb]
\centering
\caption{$\overline{\Delta}_{GEN}(\cdot)$, AlexNet trained on reg. scene imgs (5 classes). Recall: An age signal is present in $\overline{Y}$ and $\overline{Y}_{r}$ and not in $\overline{Y}_{c}$. Expectation if an age signal is learned, $\downarrow : \overline{\Delta} \approx 0$ and $\uparrow : \overline{\Delta} >> 0$. $\overline{Y}_{f}$, high frequency image components (e.g., defects) are filtered out.}
\scriptsize
\label{tab:reg_AlexNet}
\begin{tabular}{l|c|c|c|c|c}
IDs & $\overline{Y} \downarrow$ & $\overline{Y}_{c} \uparrow$ & $\overline{Y}_{r} \downarrow$ & $\overline{Y}_{f}$& $\overline{acc^{S}}$ \\ \hline
Nc01 & 0.71 & 0.66 & 0.87 & 0.75 & 0.94 \\
Nc02 & 0.74 & 0.75 & 0.80 & 0.74 & 0.94 \\
Nf01 & 0.72 & 0.72 & 0.55 & 0.72 & 0.93 \\
Nf02 & 0.73 & 0.74 & 0.73 & 0.74 & 0.94 \\
Nn01 & 0.58 & 0.68 & 0.61 & 0.67 & 0.93 \\
Nn02 & 0.68 & 0.73 & 0.76 & 0.71 & 0.93 \\
Np01 & 0.71 & 0.74 & 0.75 & 0.75 & 0.95 \\
Np02 & 0.69 & 0.72 & 0.74 & 0.77 & 0.94 \\
Ns01 & 0.70 & 0.73 & 0.80 & 0.74 & 0.96 \\
Ns02 & 0.65 & 0.70 & 0.81 & 0.74 & 0.95 \\ \hline \hline
\multicolumn{1}{c|}{$\varnothing$}& 0.69& 0.72& 0.74& 0.73& 0.94 \\
\end{tabular}
\end{table}

\begin{table*}[htb]
\centering
\caption{$\overline{\Delta}(\cdot)/\overline{\Delta}_{GEN}(\cdot)$, models trained on reg. scene imgs (2 classes). Recall: An age signal is present in $\overline{Y}$ and $\overline{Y}_{r}$ and not in $\overline{Y}_{c}$. Expectation if an age signal is learned, $\downarrow : \overline{\Delta} \approx 0$ and $\uparrow : \overline{\Delta} >> 0$. $\overline{Y}_{f}$, high frequency image components (e.g., defects) are filtered out.}
\scriptsize
\label{tab:reg_results}
\begin{tabular}{l|c|c|c|c|c || l|c|c|c|c|c }\multicolumn{6}{c}{}\\ [-0.4em]
\multicolumn{6}{c}{\textbf{SRNet}} & \multicolumn{6}{c}{\textbf{SRNet-res}} \\
IDs & $\overline{Y} \downarrow$ & $\overline{Y}_{c} \uparrow$ & $\overline{Y}_{r} \downarrow$ & $\overline{Y}_{f}$& $\overline{acc^{S}}$ & IDs & $\overline{Y} \downarrow$ & $\overline{Y}_{c} \uparrow$ & $\overline{Y}_{r} \downarrow$ & $\overline{Y}_{f}$& $\overline{acc^{S}}$ \\ \hline
Nc01 & 0.26/0.39 & 0.39/0.39 & 0.27/0.58 & 0.38/0.48 & 0.96 & Nc01 & 0.33/0.46 & 0.23/0.35 & 0.20/0.33 & 0.19/0.31 & 0.92\\
Nc02 & 0.47/0.47 & 0.41/0.41 & 0.47/0.47 & 0.47/0.47 & 0.97 & Nc02 & 0.33/0.41 & 0.29/0.49 & 0.21/0.31 & 0.28/0.36 & 0.95\\
Nf01 & 0.10/0.10 & 0.28/0.28 & 0.35/0.35 & 0.19/0.19 & 0.85 & Nf01 & 0.40/0.40 & 0.40/0.40 & 0.21/0.38 & 0.34/0.47 & 0.91\\
Nf02 & 0.27/0.27 & 0.31/0.31 & 0.26/0.26 & 0.29/0.29 & 0.82 & Nf02 & 0.40/0.41 & 0.41/0.41 & 0.21/0.30 & 0.29/0.41 & 0.91\\
Nn01 & 0.30/0.30 & 0.39/0.39 & 0.39/0.39 & 0.39/0.39 & 0.89 & Nn01 & 0.43/0.43 & 0.43/0.43 & 0.32/0.32 & 0.33/0.35 & 0.93\\
Nn02 & \textbf{-0.01/-0.01} & \textbf{0.06/0.06} & 0.36/0.36 & 0.02/0.02 & 0.86 & Nn02 & 0.45/0.45 & 0.32/0.32 & 0.14/0.15 & 0.01/0.01 & 0.95\\
Np01 & 0.44/0.44 & 0.44/0.44 & 0.21/0.21 & 0.44/0.44 & 0.94 & Np01 & 0.35/0.35 & 0.15/0.15 & 0.06/0.11 & 0.24/0.24 & 0.95\\
Np02 & 0.48/0.48 & 0.48/0.48 & 0.10/0.10 & 0.48/0.48 & 0.98 & Np02 & 0.29/0.39 & 0.24/0.55 & 0.28/0.31 & 0.32/0.38 & 0.96\\
Ns01 & 0.25/0.25 & 0.28/0.28 & 0.38/0.38 & 0.25/0.25 & 0.88 & Ns01 & 0.43/0.43 & 0.39/0.48 & 0.13/0.13 & 0.35/0.35 & 0.93\\
Ns02 & 0.34/0.34 & 0.34/0.34 & 0.31/0.31 & 0.34/0.34 & 0.84 & Ns02 & 0.37/0.37 & 0.42/0.43 & 0.33/0.33 & 0.23/0.43 & 0.93\\ 
Pn01 & 0.16/0.16 & 0.16/0.16 & 0.33/0.33 & 0.16/0.16 & 0.83 & Pn01 & 0.37/0.37 & 0.24/0.34 & 0.32/0.43 & 0.27/0.36 & 0.91\\
Pc01 & 0.18/0.20 & 0.15/0.23 & 0.25/0.25 & 0.18/0.20 & 0.75 & Pc01 & 0.41/0.41 & 0.41/0.41 & 0.29/0.29 & 0.41/0.41 & 0.91\\ \hline \hline
\multicolumn{1}{c|}{$\varnothing$}& 0.27/0.28 & 0.31/0.31 & 0.31/0.33 & 0.30/0.31 & 0.88 & \multicolumn{1}{c|}{$\varnothing$}& \textbf{0.38/0.41} & \textbf{0.33/0.40} & 0.23/0.28 & 0.27/0.34 & 0.93 \\\multicolumn{6}{c}{}\\ [-0.6em]
\multicolumn{6}{c}{\textbf{SRNet-cs}} & \multicolumn{6}{c}{\textbf{ZhuNet}} \\
IDs & $\overline{Y} \downarrow$ & $\overline{Y}_{c} \uparrow$ & $\overline{Y}_{r} \downarrow$ & $\overline{Y}_{f}$& $\overline{acc^{S}}$ & IDs & $\overline{Y} \downarrow$ & $\overline{Y}_{c} \uparrow$ & $\overline{Y}_{r} \downarrow$ & $\overline{Y}_{f}$& $\overline{acc^{S}}$ \\ \hline
Nc01   & 0.19/0.19 & 0.47/0.47 & 0.19/0.19 & 0.44/0.49 & 0.97 & Nc01 & 0.41/0.42 & 0.36/0.47 & 0.41/0.41 & 0.36/0.46 & 0.91 \\
Nc02   & 0.26/0.26 & 0.47/0.47 & 0.27/0.27 & 0.47/0.47 & 0.97 & Nc02 & 0.42/0.42 & 0.42/0.42 & 0.42/0.42 & 0.42/0.42 & 0.92 \\
Nf01   & 0.27/0.27 & 0.45/0.45 & 0.35/0.35 & 0.45/0.45 & 0.95 & Nf01 & 0.36/0.36 & 0.36/0.38 & 0.31/0.42 & 0.33/0.34 & 0.87 \\
Nf02   & 0.11/0.11 & 0.45/0.45 & 0.20/0.20 & 0.40/0.40 & 0.95 & Nf02 & 0.24/0.29 & 0.28/0.28 & 0.32/0.32 & 0.23/0.23 & 0.82 \\
Nn01   & 0.37/0.37 & 0.46/0.46 & 0.45/0.45 & 0.46/0.46 & 0.96 & Nn01 & 0.30/0.30 & 0.35/0.35 & 0.37/0.37 & 0.31/0.31 & 0.87 \\
Nn02   & 0.37/0.37 & 0.46/0.46 & 0.43/0.43 & 0.46/0.46 & 0.96 & Nn02 & 0.14/0.14 & 0.30/0.36 & 0.38/0.38 & 0.16/0.16 & 0.94 \\
Np01   & \textbf{0.06/0.06} & \textbf{0.46/0.46} & \textbf{0.07/0.07} & \textbf{0.43/0.47} & \textbf{0.96} & Np01 & 0.45/0.45 & 0.45/0.45 & 0.45/0.45 & 0.45/0.45 & 0.95 \\
Np02   & 0.20/0.20 & 0.47/0.47 & 0.11/0.11 & 0.37/0.41 & 0.97 & Np02 & 0.44/0.44 & 0.43/0.43 & 0.39/0.49 & 0.44/0.44 & 0.94 \\
Ns01   & \textbf{0.04/0.04} & \textbf{0.46/0.46} & \textbf{0.24/0.24} & \textbf{0.43/0.43} & \textbf{0.96} & Ns01 & 0.25/0.47 & 0.26/0.51 & 0.42/0.42 & 0.25/0.47 & 0.92 \\
Ns02   & 0.38/0.38 & 0.46/0.46 & 0.41/0.41 & 0.46/0.46 & 0.96 & Ns02 & 0.36/0.36 & 0.40/0.40 & 0.40/0.40 & 0.39/0.39 & 0.90 \\ 
Pn01   & 0.25/0.25 & 0.43/0.43 & 0.29/0.29 & 0.43/0.43 & 0.93 & Pn01 & 0.24/0.37 & 0.33/0.33 & 0.33/0.33 & 0.24/0.37 & 0.83 \\
Pc01   & 0.14/0.14 & 0.43/0.43 & 0.13/0.13 & 0.38/0.38 & 0.93 & Pc01 & 0.29/0.29 & 0.29/0.41 & 0.36/0.36 & 0.32/0.32 & 0.86 \\ \hline \hline
\multicolumn{1}{c|}{$\varnothing$} & 0.22/0.22 & 0.45/0.45 & 0.26/0.26 & 0.43/0.44 & 0.95 & \multicolumn{1}{c|}{$\varnothing$}& 0.33/0.36& 0.35/0.40& 0.38/0.40& 0.33/0.36& 0.89 \\
\end{tabular}
\end{table*}

The obtained accuracy values (on the test set) when using regular scene images for training are shown in Fig. \ref{fig:exp_results_reg}. In contrast to the results based on synthetic images, the best accuracy (overall) is reached when no preprocessing is applied (AlexNet and SRNet). This is reasonable if inference is based mainly on the image content.

The results for the AlexNet are reported in Table \ref{tab:reg_AlexNet}. With an average $\overline{\Delta}_{GEN}(\overline{Y})$ of 0.69 and an average $\overline{acc^{S}}$ of 0.94 (Table \ref{tab:reg_AlexNet} last row), the average prediction accuracy of $\overline{Y}$ is close to the random prediction border (\textit{i.e.}, $0.94 - 0.69 = 0.25 \approx 0.2$), indicating a strong influence of image content on the model inference. When the SRNet is trained on images of Nn02, it looks like an age signal is learned, as $\overline{\Delta}(\overline{Y})$ is even negative (Table \ref{tab:reg_results}, SRNet, bold values). However, the accuracy based on the average color ($\overline{Y}_{c}$) is similarly high. Since there is basically no age signal present in the average color, similar accuracy with the average color indicates that image content is exploited in this case as well. Across all evaluated imagers (Table \ref{tab:reg_results}, SRNet, last row) the classification accuracy is similar for all average image variants.

It is noticeable that age classification in the SRNet-res variant (Table \ref{tab:reg_results}, SRNet-res, bold values) works better with average color ($\overline{Y}_{c}$) than with average images ($\overline{Y}$). This is surprising, since the color is strongly suppressed in median filter residuals. As observed with the baseline results, $\overline{Y}_{r}$ seems to yield more stable results with the SRNet-res variant. Compared to the variants and imagers evaluated, the SRNet-cs variant when trained on images of Np01 and Ns01 (Table \ref{tab:reg_results}, SRNet-cs, bold values) seems most likely to learn an age signal (\textit{i.e.}, a low value of $\overline{\Delta}(\overline{Y})$ combined with a high value of $\overline{\Delta}(\overline{Y}_{c})$). In both cases also high-frequency parts of the average image are relevant (\textit{i.e.}, high values of $\overline{\Delta}(\overline{Y}_{f})$). This could indicate that probably in-field sensor defects are exploited. Even when looking at the results across all devices (Table \ref{tab:reg_results}, SRNet-cs, last row) with an average $\overline{\Delta}(\overline{Y})$ of 0.22 and average $\overline{acc^{S}}$ of 0.95, the average prediction accuracy of $\overline{Y}$ is clearly above the prediction by chance border (i.e., $0.95 - 0.22 = 0.73 > 0.5$), while prediction based on $\overline{Y}_{c}$ or $\overline{Y}_{f}$ does not work. This would indicate that at least part of the inference is based on an age signal. In contrast, considering the ZhuNet (Table \ref{tab:reg_results}, ZhuNet), the overall age classification accuracy of $\overline{Y}$ is close to the random prediction, indicating a strong influence of the image content.

When comparing $\overline{\Delta}(\cdot)$ and $\overline{\Delta}_{GEN}(\cdot)$ a significant difference can only be observed with the model SRNet-res trained on regular scene images. In total, 11\% of all evaluated runs (4 average image variants $\times$ 12 imagers $\times$ 8 runs) have an accuracy of less than 0.5. This is also significantly higher than for SRNet, SRNet-cs and ZhuNet with values of 3\%, 2\% and 5\% respectively. When the models are trained on synthetic images (robust detection of an age signal), no accuracy values of less than 0.5 are observed.

\section{Conclusion}
\label{sec:con}
Content bias is an important problem in the context of deep learning image age approximation. Based on the experiments conducted, we have seen that it is difficult for a `standard' CNN to exploit the subtle age traces, even if it is not distracted by content bias in the training data. In this work, we proposed a method that evaluates the influence of image content (based on different types of average images). The functionality of this method was validated using models trained on synthetic images (where content bias can be ruled out). Based on this approach, we showed that `standard' CNNs (AlexNet and SRNet) trained on regular images are strongly dependent on image content. Image content still plays an important role, even when pre-processing is applied to increase the signal-to-noise ratio (\textit{i.e.}, SRNet-res, SRNet-cs and ZhuNet). Based on the results, image content is least involved in inference with the SRNet-cs. However, since image content is still involved, none of the investigated variants is considered suitable for image age approximation. Thus, we recommend classical image age approximation methods \cite{Fridrich11a,joechl20a} at current stage.

\bibliographystyle{elsarticle-num}
\bibliography{bib}
\end{document}